%% file: emnlp2021.tex
\pdfoutput=1

\documentclass[11pt]{article}

\usepackage[]{emnlp2021}

\usepackage{times}
\usepackage{latexsym}
\usepackage{multirow}
\usepackage{graphicx}
\usepackage{booktabs}
\usepackage{mathtools}
\usepackage{xspace}
\usepackage{arydshln}
\newcommand{\eg}{{\em e.g.,}\xspace}

\usepackage[nameinlink]{cleveref}

\crefformat{section}{\S#2#1#3} 
\crefname{algorithm}{Alg.}{Algs.}
\crefformat{subsection}{\S#2#1#3}
\Crefname{equation}{Eq.}{Eqs.}
\Crefname{figure}{Fig.}{Figs.}

\usepackage[T1]{fontenc}

\usepackage[utf8]{inputenc}

\usepackage{microtype}
\usepackage{booktabs} 

\usepackage{adjustbox}
\usepackage[frozencache,cachedir=.]{minted}
\usepackage{siunitx}
\definecolor{pastel_blue}{rgb}{0.86, 0.926, 0.984}

\title{CodeT5: Identifier-aware Unified Pre-trained Encoder-Decoder Models for Code Understanding and Generation}

\author{Yue Wang$^{1}$,~Weishi Wang$^{12}$,~Shafiq Joty$^{12}$,~and Steven C.H. Hoi$^{1}$\\
$^{1}$ Salesforce Research Asia \\
$^{2}$ Nanyang Technological University, Singapore\\
\texttt{\{wang.y,weishi.wang,sjoty,shoi\}@salesforce.com}\\}

\begin{document}
\maketitle
\input{sections/abstract}
\input{sections/intro}
\input{sections/related}
\input{sections/model}
\input{sections/exp-setup}
\input{sections/exp-res}

\input{sections/conclusion}

\input{sections/impact}
\input{sections/ack}
\bibliography{emnlp2021}
\bibliographystyle{acl_natbib}

\end{document}

%% file: sections/abstract.tex
\begin{abstract}
Pre-trained models for Natural Languages (NL) like BERT and GPT have been recently shown to transfer well to Programming Languages (PL) and largely benefit a broad set of code-related tasks. Despite their success, most current methods either rely on an encoder-only (or decoder-only) pre-training that is suboptimal for generation (resp. understanding) tasks or process the code snippet in the same way as NL, neglecting the special characteristics of PL such as token types. We present CodeT5, a unified pre-trained encoder-decoder Transformer model that better leverages the code semantics conveyed from the developer-assigned identifiers. Our model employs a unified framework to seamlessly support both code understanding and generation tasks and allows for multi-task learning. Besides, we propose a novel identifier-aware pre-training task that enables the model to distinguish which code tokens are identifiers and to recover them when they are masked. Furthermore, we propose to exploit the user-written code comments with a bimodal dual generation task for better NL-PL alignment.
Comprehensive experiments show that CodeT5 significantly outperforms prior methods on understanding tasks such as code defect detection and clone detection, and generation tasks across various directions including PL-NL, NL-PL, and PL-PL. Further analysis reveals that our model can better capture semantic information from code.  Our code and pre-trained models are released at \url{https://github.com/salesforce/CodeT5}.
\end{abstract}

%% file: sections/intro.tex
\section{Introduction}
Pre-trained language models such as 
BERT~\cite{DBLP:conf/naacl/DevlinCLT19}, GPT~\cite{radford2019language}, and T5~\cite{DBLP:journals/jmlr/RaffelSRLNMZLL20} have greatly boosted  performance  in a wide spectrum of natural language processing (NLP) tasks. They typically employ a pre-train then fine-tune paradigm that aims to derive generic language representations by self-supervised training on large-scale unlabeled data, which can be transferred to benefit multiple downstream tasks, especially those with limited data annotation.
Inspired by their success, there are many recent attempts to adapt these pre-training methods for programming language (PL)~\cite{DBLP:conf/sigsoft/SvyatkovskiyDFS20,DBLP:conf/icml/KanadeMBS20,DBLP:conf/emnlp/FengGTDFGS0LJZ20}, showing promising results on  code-related tasks.

\input{figures/finetune_tasks}

However, despite their success, most of these models rely on either an encoder-only model similar to BERT~\cite{DBLP:conf/sigsoft/SvyatkovskiyDFS20,DBLP:conf/emnlp/FengGTDFGS0LJZ20} or a decoder-only model like GPT~\cite{DBLP:conf/icml/KanadeMBS20}, which is suboptimal for generation and understanding tasks, respectively. For example, CodeBERT~\cite{DBLP:conf/emnlp/FengGTDFGS0LJZ20} requires an additional decoder when applied for the code summarization task, where this decoder cannot benefit from the pre-training. 
Besides, most existing methods simply employ the conventional NLP pre-training techniques on source code by regarding it as a sequence of tokens like NL. This largely ignores the rich structural information in code, which is vital to fully comprehend the  code semantics.

In this work, we present CodeT5, a pre-trained encoder-decoder model  that  considers the token type information in code.
Our CodeT5 builds on the T5 architecture~\cite{DBLP:journals/jmlr/RaffelSRLNMZLL20} that employs denoising sequence-to-sequence (Seq2Seq) pre-training and has been shown to benefit both understanding and generation tasks in natural language.
In addition, we propose to leverage the  developer-assigned identifiers in code.
When writing programs, developers tend to employ informative identifiers to make the code more understandable, so that these identifiers would generally preserve rich code semantics, \eg the ``binarySearch'' identifier in Figure~\ref{fig:pretrain_task} directly indicates its functionality. To fuse such code-specific knowledge, we propose a novel identifier-aware objective that trains the model to  distinguish which tokens are identifiers and recover them when they are masked.

Furthermore, we propose to leverage the code and its accompanying comments to learn a better NL-PL alignment. Developers often provide documentation for programs to facilitate better software maintenance~\cite{DBLP:conf/sigdoc/SouzaAO05}, so that such PL-NL pairs are widely available in most source code. Specifically, we regard the NL$\rightarrow$PL generation and   PL$\rightarrow$NL generation as dual tasks and simultaneously optimize the model on them. 

We pre-train CodeT5 on the CodeSearchNet data~\cite{DBLP:journals/corr/abs-1909-09436} following~\cite{DBLP:conf/emnlp/FengGTDFGS0LJZ20} that consists of both unimodal (PL-only) and bimodal (PL-NL) data on six PLs. 
In addition to that, we further collect extra data of C/C\# from open-source Github repositories. 
We fine-tune CodeT5 on most tasks in the CodeXGLUE benchmark~\cite{DBLP:journals/corr/abs-2102-04664}, including two understanding tasks: code defect detection and clone detection, and generation tasks such as code summarization,  generation, translation, and refinement. 
As shown in Figure~\ref{fig:finetune_task},
we also explore multi-task learning to fine-tune CodeT5 on multiple tasks at a time using a task control code as the source prompt.

\noindent In summary, we make the following contributions: 
\begin{itemize}
\vspace{-0.5em}
\itemsep0em
    \item We present one of the first unified encoder-decoder models CodeT5 to support both code-related understanding and generation tasks, and also   allows for multi-task learning. 
    \item We propose a novel identifier-aware pre-training objective that considers the crucial token type information (identifiers) from code. Besides, we propose to leverage the NL-PL pairs that are naturally available in source code to learn a better cross-modal alignment.
    \item 
    Extensive experiments show that CodeT5 yields state-of-the-art results on the fourteen sub-tasks in CodeXGLUE. Further analysis shows our CodeT5 can better capture the code semantics with the proposed identifier-aware pre-training and bimodal dual generation primarily benefits NL$\leftrightarrow$PL tasks.
\end{itemize}

%% file: figures/finetune_tasks.tex
\begin{figure*}[t]
\centering
\includegraphics[width=0.85\textwidth]{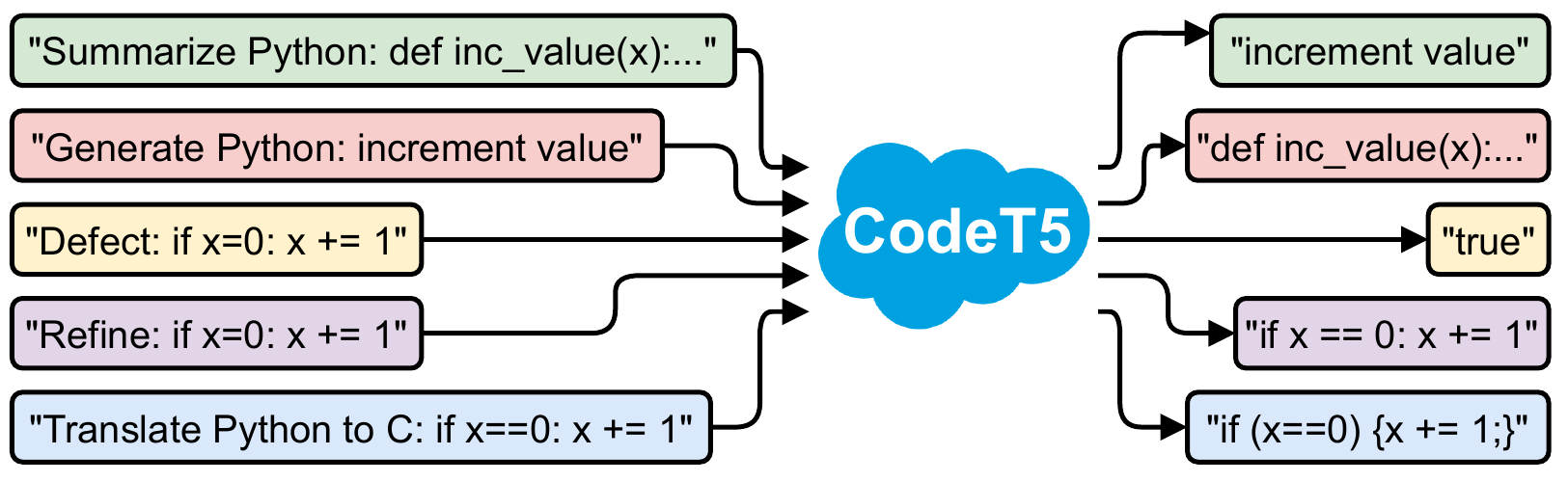}
\vspace{-0.5em}
\caption{Illustration of our CodeT5 for code-related understanding and generation  tasks.}\label{fig:finetune_task}
\vspace{-1em}
\end{figure*}



%% file: sections/related.tex
\section{Related Work}

\paragraph{Pre-training on Natural Language.}
Pre-trained models based on Transformer architectures~\cite{DBLP:conf/nips/VaswaniSPUJGKP17} have led to state-of-the-art performance on a broad set of NLP tasks.
They can be generally categorized into three groups: encoder-only models such as BERT~\cite{DBLP:conf/naacl/DevlinCLT19}, RoBERTa~\cite{DBLP:journals/corr/abs-1907-11692}, and ELECTRA~\cite{DBLP:conf/iclr/ClarkLLM20}, decoder-only models like GPT~\cite{radford2019language}, and encoder-decoder models such as MASS~\cite{DBLP:conf/icml/SongTQLL19}, BART~\cite{DBLP:conf/acl/LewisLGGMLSZ20}, and T5~\cite{DBLP:journals/jmlr/RaffelSRLNMZLL20}.
Compared to encoder-only and decoder-only models that respectively favor understanding and generation tasks, encoder-decoder models can well support both types of tasks. They often employ denoising sequence-to-sequence pre-training objectives that corrupt the source input and require the decoder to recover them. 
In this work, we extend T5 to the programming language and propose a novel identifier-aware denoising objective that enables the model to better comprehend the code.

\input{figures/pretrain_tasks}

\paragraph{Pre-training on Programming Language.}

Pre-training on the programming language is a nascent field where much recent work attempts to extend the NLP pre-training methods to source code. CuBERT~\cite{DBLP:conf/icml/KanadeMBS20} and CodeBERT~\cite{DBLP:conf/emnlp/FengGTDFGS0LJZ20}  are  the two pioneer models. CuBERT employs BERT's powerful masked language modeling objective to derive generic code-specific representation, and CodeBERT further adds a replaced token detection~\cite{DBLP:conf/iclr/ClarkLLM20} task to learn NL-PL cross-modal representation.
Besides the BERT-style models, \citet{DBLP:conf/sigsoft/SvyatkovskiyDFS20} and \citet{DBLP:conf/kbse/LiuLZJ20} respectively employ GPT and UniLM~\cite{DBLP:conf/nips/00040WWLWGZH19} for the code completion task.
Transcoder~\cite{DBLP:conf/nips/RoziereLCL20} explores programming language translation in an unsupervised setting.
Different from them, we explore encoder-decoder models based on T5 for programming language pre-training and support a more comprehensive set of tasks.

Some emerging work~\cite{DBLP:conf/emnlp/ClementDTSS20,DBLP:conf/icse/MastropaoloSCNP21,DBLP:journals/corr/abs-2104-02443} in  the recent literature also  explore the T5 framework on code, but they only focus on a limited subset of generation tasks and do not support understanding tasks like us.
Apart from these, 
PLBART~\cite{DBLP:journals/corr/abs-2103-06333} based on another encoder-decoder model BART  can also  support both understanding and generation tasks. However, all the above prior work simply processes code in the same way as natural language and largely ignores the code-specific characteristics.
Instead, we propose to leverage the  identifier information in code for pre-training.

Recently,  GraphCodeBERT~\cite{DBLP:journals/corr/abs-2009-08366} incorporates the  data flow extracted from the code structure into CodeBERT,  while \citet{DBLP:journals/corr/abs-2102-07492} propose a deobfuscation objective to leverage the structural aspect of PL. These models only focus on training a better code-specific encoder. 
\citet{DBLP:conf/iclr/ZugnerKCLG21} proposes to capture the relative distances between code tokens over the code structure.
By contrast, we specifically focus on the identifiers that reserve rich code semantics and fuse such information into a Seq2Seq model via two novel identifier tagging and prediction tasks.

%% file: figures/pretrain_tasks.tex
\begin{figure*}
\centering
\includegraphics[trim={0cm 0.2cm 0 1cm}, scale=0.52]{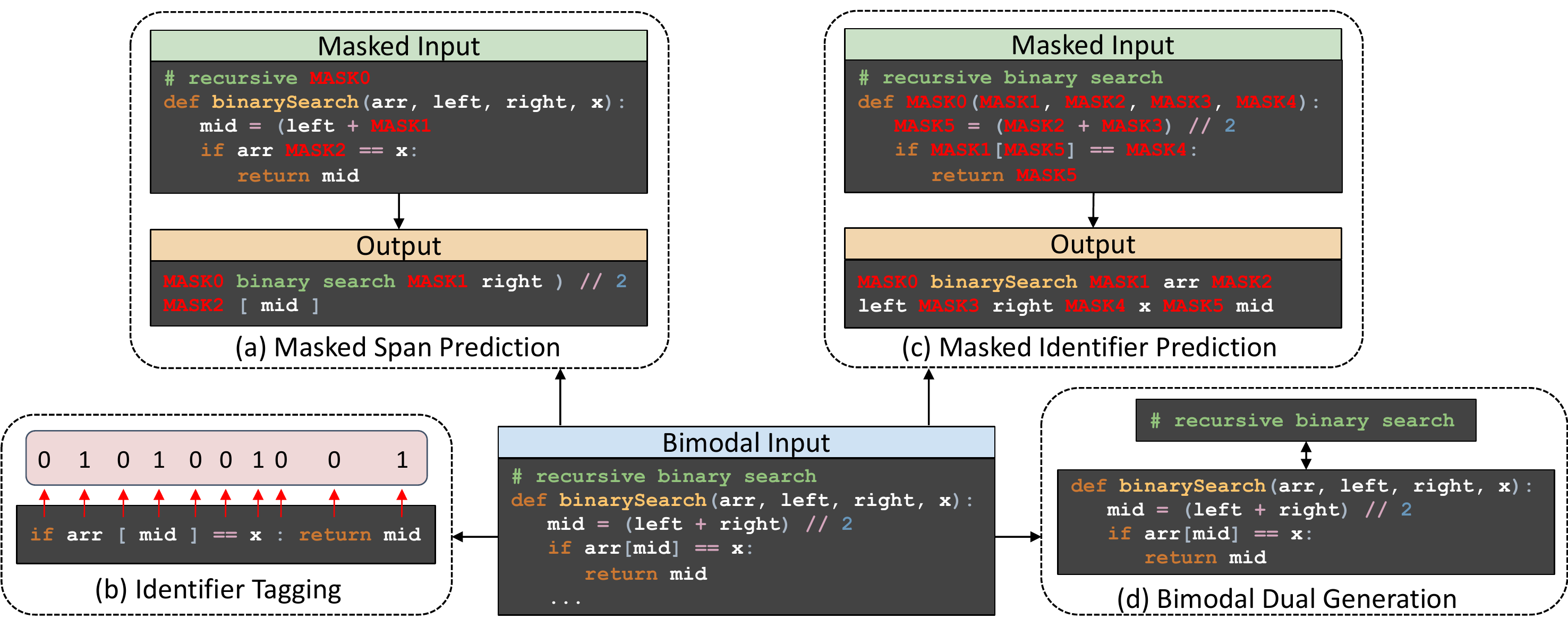}
\vspace{-0.3em}
\caption{Pre-training tasks of CodeT5. We first alternately train span prediction, identifier prediction, and identifier tagging on both unimodal  and bimodal data, and then leverage the bimodal data for dual generation training. }\label{fig:pretrain_task}
\vspace{-0.8em}
\end{figure*}

%% file: sections/model.tex
\section{CodeT5}

Our CodeT5 builds on an encoder-decoder framework with the same architecture as T5~\cite{DBLP:journals/jmlr/RaffelSRLNMZLL20}.
It aims to derive generic representations for programming language (PL) and natural language (NL)  via pre-training on unlabeled source code.
As illustrated in Figure~\ref{fig:pretrain_task},
we extend the denoising  Seq2Seq objective in T5 by proposing two 
identifier tagging and prediction tasks to enable the model to better leverage the  token type information from PL, which are the identifiers assigned by developers.
To improve the NL-PL alignment, we further propose a bimodal dual learning objective for a bidirectional conversion between NL and PL.

In the following, we introduce how CodeT5 encodes PL and NL inputs~(\cref{model:input}) and our proposed identifier-aware pre-training tasks (\cref{model:pretrain}), followed by the fine-tuning with task-specific transfer learning and multi-task training~(\cref{model:finetune}).

\subsection{Encoding NL and PL}\label{model:input}
At the pre-training stage, our model would receive either PL-only or NL-PL as inputs depending on whether the code snippet has accompanying NL descriptions or not. For the NL-PL bimodal inputs, we concatenate them into a sequence with a delimiter token  \texttt{[SEP]} and represent the whole input sequence into the format as $\mathbf{x}$ $=$ (\texttt{[CLS]}, $w_1,...,w_n$, \texttt{[SEP]}, $c_1,...,c_m$, \texttt{[SEP]}), where $n$ and $m$ denote the number of NL word tokens  and PL code tokens, respectively. The NL word sequence will be empty for PL-only unimodal inputs. 

In order to capture more code-specific features, we propose to leverage token type information from code. We focus on the type of identifiers (\eg function names and variables) as they are one of the most PL-agnostic features and reserve rich code semantics.
Specifically, we convert the PL segment into an Abstract Syntax Tree (AST) and extract the node types for each code token. Finally, we construct a sequence of binary labels $\mathbf{y} \in {\{0,1\}^m}$ for the PL segment, where each $y_i\in \{0, 1\}$ represents whether the code token $c_i$ is an identifier or not.

\subsection{Pre-training Tasks}\label{model:pretrain}

We now introduce our proposed pre-training tasks that enable CodeT5 to learn useful patterns from either PL-only or NL-PL bimodal data. 

\paragraph{Identifier-aware Denoising Pre-training.}
Denoising Sequence-to-Sequence (Seq2Seq) pre-training has been shown to be quite effective in a broad set of NLP tasks~\cite{DBLP:conf/icml/SongTQLL19,DBLP:journals/jmlr/RaffelSRLNMZLL20,DBLP:conf/acl/LewisLGGMLSZ20}. This denoising objective typically first corrupts the source sequence with some noising functions and then requires the decoder to recover the original texts.
In this work, we utilize a  span masking objective similar to T5~\cite{DBLP:journals/jmlr/RaffelSRLNMZLL20} that randomly masks spans with arbitrary lengths and then predicts these masked spans combined with some sentinel tokens at the decoder.
We refer this task to \textbf{Masked Span Prediction (MSP)}, as illustrated in Figure~\ref{fig:pretrain_task} (a).

Specifically, we employ the same $15\%$ corruption rate as T5 and ensure the average span length to be $3$ by uniformly sampling spans of from $1$ to $5$ tokens. Moreover, we employ the \emph{whole word masking} by sampling spans  before subword tokenization, which aims to avoid masking partial sub-tokens and is shown to be helpful~\cite{DBLP:journals/corr/abs-1904-09223}. 
Notably, we pre-train a shared model for various PLs to learn robust  cross-lingual representations. 
We  describe  the masked span prediction loss as:
\begin{equation}\label{eq:S2S}
\vspace{-0.5em}
 \small 
\hspace{-0.8em}  \mathcal{L}_{MSP} (\theta) = \sum_{t = 1}^{k} \hspace{-0.5em} -\log P_{\theta} (x^{\text{mask}}_t| \mathbf{x}^{\backslash \text{mask}}, \mathbf{x}^{\text{mask}}_{<t}),
\normalsize  
\end{equation}

\noindent where $\theta$ are the model parameters, $\mathbf{x}^{\backslash \text{mask}}$ is the masked input, $\mathbf{x}^{\text{mask}}$ is the masked sequence to predict from the decoder with $k$ denoting the number of tokens in $\mathbf{x}^{\text{mask}}$, and $\mathbf{x}^{\text{mask}}_{<t}$ is the span sequence generated so far.
 
 To fuse more code-specific structural information (the identifier node type in AST) into the model, we propose two additional tasks: \textit{Identifier Tagging (IT)} and \textit{Masked Identifier Prediction (MIP)} to complement the denoising pre-training.
.
 
\paragraph{$\bullet$~Identifier Tagging (IT)} 
It aims to notify the model with the knowledge of whether this code token is an identifier or not, which shares a similar spirit of syntax highlighting in some developer-aided tools.
As shown in Figure~\ref{fig:pretrain_task} (b), we map the final hidden states of the PL segment at the CodeT5 encoder into a sequence of probabilities $\mathbf{p}=(p_1, ...,p_m)$, and compute a binary cross entropy loss for sequence labeling:
 \begin{equation}\label{eq:it}
 \small 
 \vspace{-0.3em}
 \mathcal{L}_{IT} (\theta_{e}) = \sum_{i=1}^m \hspace{-0.2em} -[y_i \log p_i + (1-y_i) \log (1-p_i)],
 \normalsize
 \end{equation}
\noindent where $\theta_e$ are the encoder parameters. Note that by casting the task as a sequence labeling problem, the model is expected to capture the code syntax and the data flow structures of the code.
 
\paragraph{$\bullet$~Masked Identifier Prediction (MIP)} 
Different from the random span masking in MSP, we mask all identifiers in the PL segment and employ a unique sentinel token for all occurrences of one specific identifier. In the field of software engineering, this is called \emph{obfuscation} where changing identifier names does not impact the code semantics.  Inspired by~\citet{DBLP:journals/corr/abs-2102-07492}, we arrange the unique identifiers with the sentinel tokens into  a target sequence $\mathbf{I}$ as shown in Figure~\ref{fig:pretrain_task} (c).
We then predict it in an auto-regressive manner:
\begin{equation}\label{eq:IP}
\small 
\vspace{-0.5em}
  \mathcal{L}_{MIP} (\theta) = \sum_{j = 1}^{|I|} -\log P_{\theta}(I_j| \mathbf{x}^{\backslash \mathbf{I}}, \mathbf{I}_{<j}),
  \normalsize
  \vspace{-0.5em}
\end{equation}
\noindent where $\mathbf{x}^{\backslash \mathbf{I}}$ is the masked input. Note that \emph{deobfuscation} is a more challenging task that requires the model to comprehend the code semantics based on obfuscated code and link the occurrences of the same identifiers together.

We alternately optimize these three losses with an equal probability, which constitutes our proposed identifier-aware  denoising  pre-training.

\paragraph{Bimodal Dual Generation.}
In the  pre-training phase, the decoder only sees  discrete masked spans and identifiers, which is disparate from the downstream tasks where the decoder needs to generate either fluent NL texts or syntactically correct code snippets. To close the gap between the pre-training and fine-tuning, we propose to leverage the NL-PL bimodal data to train the model for a bidirectional conversion as shown in Figure~\ref{fig:pretrain_task}~(d).
Specifically, we regard the NL$\rightarrow$PL generation and   PL$\rightarrow$NL generation as  dual tasks and  simultaneously optimize the model on them.
For each NL-PL bimodal datapoint, we construct two training instances with reverse directions and add language ids (\eg <java> and <en> for Java PL and English NL, respectively). This operation can be also seen as a special case of T5's span masking by either masking the full NL or PL segment from the bimodal inputs. 
This task aims to improve the alignment between the NL and PL counterparts.

\subsection{Fine-tuning CodeT5} \label{model:finetune}
After pre-training on large-scale unlabeled data, we adapt CodeT5 to downstream tasks via either task-specific transfer learning or multi-task learning.

\paragraph{Task-specific Transfer Learning: Generation vs. Understanding Tasks.} Code-related tasks can be categorized into generation and understanding tasks. For the former one, our CodeT5 can  be naturally adapted with its Seq2Seq framework. 
For understanding tasks, we investigate two ways of either generating the label as a unigram target sequence \cite{DBLP:journals/jmlr/RaffelSRLNMZLL20}, or predicting it from the vocabulary of class labels based on the last decoder hidden state following~\citet{DBLP:conf/acl/LewisLGGMLSZ20}.

\paragraph{Multi-task Learning.}
We also explore a multi-task learning setting by training a shared model on multiple tasks at a time. 
Multi-task learning is able to reduce computation cost by reusing the most of model weights for many tasks and has been shown to improve the model generalization capability in NL pre-training~\cite{DBLP:conf/acl/LiuHCG19}.
We follow~\citet{DBLP:journals/jmlr/RaffelSRLNMZLL20} to employ the same unified model for all tasks without adding any task-specific networks but allow to select different best checkpoints for different tasks. 
To notify the model with which task it is dealing with, we design a unified format of task control codes and prepend it into the source inputs as shown in Figure~\ref{fig:finetune_task}. For instance, we employ ``Translate Java to CSharp:'' as the source prompt for the code-to-code translation task from Java to CSharp.

As different tasks have different dataset sizes, we follow~\citet{DBLP:conf/nips/ConneauL19} to employ a balanced sampling strategy. For $N$ number of datasets (or tasks), with probabilities ${\{q_i\}}_{i=1}^{N}$, we define the following multinomial distribution to sample from:
\begin{equation}\label{eq:sample}
q_i=\frac{r^\alpha_i}{\sum^{N}_{j=1}r^\alpha_j} \text{, where }   \ \ r_i=\frac{n_i}{\sum^{N}_{k=1}n_k} \text{,}
\end{equation}
\noindent where $n_i$ is number of examples for $i$-th task and $\alpha$ is set to $0.7$. This balanced sampling aims to alleviate the bias towards high-resource tasks.

%% file: sections/exp-setup.tex
\section{Experimental Setup}
\input{tables/dataset}

\subsection{Pre-training Dataset}
We follow~\citet{DBLP:conf/emnlp/FengGTDFGS0LJZ20} to employ CodeSearchNet~\cite{DBLP:journals/corr/abs-1909-09436} to pre-train CodeT5, which consists of six PLs with both unimodal and bimodal data.  Apart from that, we additionally collect two datasets of C/CSharp from BigQuery\footnote{\url{https://console.cloud.google.com/marketplace/details/github/github-repos}} to ensure that all downstream tasks have overlapped PLs with the pre-training data. In total, we employ around 8.35 million instances for pre-training. Table~\ref{table:data} shows some basic statistics.
To obtain the identifier labels from code, we leverage the tree-sitter\footnote{\url{https://tree-sitter.github.io/tree-sitter/}} to convert the PL into an abstract syntax tree  and then extract its node type information. 
We filter out reserved keywords for each PL from its identifier list. 
We observe that PLs have different identifier rates, where  Go has the least rate of $19\%$ and  Ruby has the highest rate of $32\%$.

\subsection{Code-specific Tokenizer}
Tokenization is a key ingredient for the success of pre-trained language models like BERT and GPT. They often employ a Byte-Pair Encoding (BPE) tokenizer~\cite{DBLP:conf/acl/SennrichHB16a} to alleviate the Out-of-Vocabulary (OoV) issues.
Specifically, we train a Byte-level BPE tokenizer following~\citet{radford2019language} and  set the vocabulary size to 32,000 as T5. We add additional special tokens (\texttt{[PAD]}, \texttt{[CLS]}, \texttt{[SEP]}, \texttt{[MASK0]}, ..., \texttt{[MASK99]}). 
This tokenzier is trained on all of our pre-training data with non-printable characters and low-frequent tokens (occurring <3 times) filtered. We compare it with T5's default tokenizer and find that our tokenizer largely reduces the length of tokenized code sequence by  $30\%$ - $45\%$ on downstream tasks. 
This will   accelerate the training and especially  benefit generation tasks due to the shorter sequence to predict.
We also spot a severe problem for applying the T5's default tokenizer on source code, where it would encode some common code tokens such as brackets [`\{', `\}'] into unknown tokens.

\input{tables/NL_PL_gen}

\subsection{Downstream Tasks and  Metrics}
We cover most generation and understanding tasks in the CodeXGLUE benchmark~\cite{DBLP:journals/corr/abs-2102-04664} and employ the provided public datasets and the same data splits following it for all these tasks. 

We first consider two cross-modal generation tasks. \textbf{Code summarization} aims to summarize a function-level code snippet into English descriptions. The dataset consists of six PLs including Ruby, JavaScript, Go, Python, Java, and PHP from CodeSearchNet~\cite{DBLP:journals/corr/abs-1909-09436}.  We employ the smoothed BLEU-4~\cite{DBLP:conf/coling/LinO04} to evaluate  this task.
\textbf{Code generation} is the task to generate a code snippet based on NL descriptions.  We employ the Concode data~\cite{DBLP:conf/emnlp/IyerKCZ18} in Java where the input contains both NL texts and class environment contexts, and the output is a function. We evaluate it with BLEU-4, exact match (EM) accuracy, and CodeBLEU~\cite{DBLP:journals/corr/abs-2009-10297} that considers  syntactic and semantic matches based on the code structure in addition to the n-gram match.

Besides, we consider two code-to-code generation tasks. \textbf{Code translation} aims to migrate legacy software from one PL to another, where we focus on translating functions from Java to CSharp and vice versa. \textbf{Code refinement} aims to convert a buggy function into a correct one. We employ two Java datasets provided by~\citet{DBLP:journals/tosem/TufanoWBPWP19} with various function lengths: small (fewer than 50 tokens) and medium (50-100 tokens). 
We use BLEU-4 and exact match to evaluate them.

We also investigate how CodeT5 performs  on  two understanding-based tasks. 
The first one is \textbf{defect detection} that aims to predict whether a code is vulnerable to software systems or not. We use the C dataset provided by \citet{DBLP:conf/nips/ZhouLSD019} for experiment.
The second task is \textbf{clone detection} which aims to measure the similarity between two code snippets and predict whether they have the same functionality. We experiment with the Java data provided by \citet{DBLP:conf/wcre/WangLM0J20}. We employ F1 score and accuracy for evaluating these two tasks respectively. 
In total, our CodeT5 supports six tasks and fourteen sub-tasks in CodeXGLUE with a unified encoder-decoder model.

\subsection{Comparison Models}
We compare CodeT5 with state-of-the-art (SOTA) pre-trained models that can be categorized into three types:  encoder-only, decoder-only, and encoder-decoder models.
As \textbf{encoder-only} models, we consider RoBERTa~\cite{DBLP:journals/corr/abs-1907-11692}, RoBERTa (code) trained with masked language modeling (MLM) on code, CodeBERT~\cite{DBLP:conf/emnlp/FengGTDFGS0LJZ20} trained with both MLM and replaced token detection~\cite{DBLP:conf/iclr/ClarkLLM20},  GraphCodeBERT~\cite{DBLP:journals/corr/abs-2009-08366} using data flow from code, and DOBF~\cite{DBLP:journals/corr/abs-2102-07492} trained with the identifier deobfuscation objective. Note that although DOBF employs a Seq2Seq model during pre-training, it only aims to train a better encoder for downstream tasks without exploring the potential benefit of the pre-trained decoder. 

For \textbf{decoder-only} models, we compare GPT-2~\cite{radford2019language} and its adaptations on code domain including CodeGPT-2, and CodeGPT-adapted. The difference is that the latter one utilizes a GPT-2 checkpoint for model initialization while the former one is trained from scratch. As \textbf{encoder-decoder} models, the current SOTA model for the CodeXGLUE benchmark is PLBART~\cite{DBLP:journals/corr/abs-2103-06333} based on BART~\cite{DBLP:conf/acl/LewisLGGMLSZ20} architecture.
For pre-training data, most of these models employ CodeSearchNet~\cite{DBLP:journals/corr/abs-1909-09436} except  DOBF and PLBART.
DOBF is pre-trained on  7.9M Java and 3.6M Python files from BigQuery while PLBART employs a much larger data   with 470M Python and 210M Java functions, and  47M NL posts from StackOverflow.

\input{tables/PL_PL_understand}

\subsection{Model Configurations}
We build CodeT5 based on Huggingface's  T5~\cite{DBLP:journals/jmlr/RaffelSRLNMZLL20} PyTorch implementation\footnote{\url{https://huggingface.co/}} and employ two sizes of CodeT5-small (60M) and CodeT5-base (220M).
We set the maximum source and target sequence lengths to be $512$ and $256$, respectively.
We use the mixed precision of FP16 to accelerate the pre-training. We set the batch size to $1024$ and employ the peak learning rate of 2e-4 with  linear decay.
We pre-train the model with the denoising objective for $100$ epochs and  bimodal dual training  for further $50$ epochs on a cluster of $16$ NVIDIA A100 GPUs with $40$G memory. 
The total training time for CodeT5-small and CodeT5-base is $5$ and $12$ days, respectively.

In the fine-tuning phase, we find that the tasks in  CodeXGLUE~\cite{DBLP:journals/corr/abs-2102-04664}  are quite sensitive to  some hyper parameters such as learning rate, training steps, and batch size. We conduct a grid search  and select the best parameters based on the validation set.
In  multi-task learning, we cover all downstream tasks except  clone detection. 

%% file: tables/dataset.tex
\begin{table}[!t]
\centering
\resizebox{0.85\linewidth}{!}{%

\begin{tabular}{p{0.02\textwidth}l|c c c}
\toprule
& PLs & W/ NL  & W/o NL & Identifier   \\
\midrule
\multirow{6}{*}{\parbox[t]{2mm}{\rotatebox[origin=c]{90}{CodeSearchNet}} $\begin{dcases*} \\ \\ \\ \\ \\  \end{dcases*}$}  
&Ruby &49,009 & 110,551 & 32.08\% \\
&JavaScript &125,166 & 1,717,933 & 19.82\%  \\
&Go  & 319,132 & 379,103 & 19.32\% \\
&Python &453,772& 657,030 & 30.02\% \\ 
&Java &457,381& 1,070,271 & 25.76\%  \\
&PHP &525,357& 398,058 & 23.44\%  \\
\multirow{2}{*}{\parbox[t]{2mm}{\rotatebox[origin=c]{90}{Our}} $\begin{dcases*} \\   \end{dcases*}$}  
&C &1M& - & 24.94\% \\
&CSharp &228,496 & 856,375 & 27.85\% \\
\midrule
&Total & 3,158,313 & 	5,189,321  & 8,347,634  \\
\bottomrule
\end{tabular}
}
\vspace{-0.5em}
\caption{Dataset statistics.   ``Identifier'' denotes the proportion of identifiers over all code tokens for each PL. \label{table:data}
}
\vspace{-2em}
\end{table}

%% file: tables/NL_PL_gen.tex
\begin{table*} [t]
\begin{center}
\resizebox{0.99\textwidth}{!}{
\begin{tabular}{p{12cm}p{7cm}}
\begin{minipage}{0.75\textwidth}
\resizebox{\textwidth}{!}{
\begin{tabular}{lc c c c c cc}
\toprule
Methods & Ruby & JavaScript & Go & Python & Java & PHP & Overall \\
\midrule
RoBERTa  & 11.17 & 11.90 & 17.72 & 18.14 & 16.47 & 24.02 & 16.57 \\
CodeBERT & 12.16 & 14.90 & 18.07 & 19.06 & 17.65 & 25.16 & 17.83 \\ 
DOBF & -& -& -  &18.24 & 19.05 & - & - \\
PLBART &  14.11 &  15.56 &  18.91 & 19.30 & 18.45 & 23.58 & 18.32\\

\midrule
CodeT5-small & 14.87&	15.32&	19.25&	20.04&	19.92&	25.46&	19.14\\
~~~+dual-gen & 15.30 &	15.61&	19.74&	19.94&	19.78&	26.48&	19.48 \\
~~~+multi-task & 15.50&	15.52&	19.62&	20.10&	19.59&	25.69&	19.37 \\
\cdashline{1-8}
CodeT5-base & 15.24&	16.16&	19.56&	20.01&	20.31&	26.03&	19.55\\
~~~+dual-gen & \textbf{15.73}&	16.00&	19.71&	20.11&	20.41&	\textbf{26.53}&	19.75\\
~~~+multi-task & 
15.69&	\textbf{16.24}&	\textbf{19.76}&	\textbf{20.36}&	\textbf{20.46}&	26.09&	\textbf{19.77}\\
\bottomrule
\end{tabular}
}
\captionsetup{type=table}
\vspace{-0.5em}  
\caption{Smoothed BLEU-4 scores on the code summarization task.
The ``Overall'' column shows the average scores over six PLs. Best results are in bold.
}\label{table:summarize}

\end{minipage}

&
\begin{minipage}{0.48\textwidth}
\resizebox{\textwidth}{!}{
\begin{tabular}{lc c c}
\toprule
Methods & EM & BLEU & CodeBLEU \\ 
\midrule
GPT-2 & 17.35 & 25.37 & 29.69 \\
CodeGPT-2 & 18.25 & 28.69 & 32.71 \\
CodeGPT-adapted & 20.10 & 32.79 & 35.98 \\ 
PLBART & 18.75 &  36.69 & 38.52 \\
\midrule

CodeT5-small &21.55	&38.13	&41.39 \\
~~~+dual-gen	&19.95	&39.02	&42.21 \\
~~~+multi-task	& 20.15 &35.89 &38.83 \\
\cdashline{1-4}

CodeT5-base &22.30 &40.73 &43.20 \\
~~~+dual-gen	& \textbf{22.70} &	\textbf{41.48}&	\textbf{44.10}\\
~~~+multi-task	& 21.15 & 37.54 & 40.01 \\

\bottomrule
\end{tabular}
}
\captionsetup{type=table}
\vspace{-0.5em}  
\caption{Results on the code generation task.
EM denotes the exact match.
}
\label{table:concode}

\end{minipage}
\end{tabular}
}
\end{center}
\vspace{-2em}
\end{table*}

%% file: tables/PL_PL_understand.tex
\begin{table*} [t]
\begin{center}
\resizebox{\textwidth}{!}{
\begin{tabular}{p{12cm}p{6cm}}
\begin{minipage}{0.76\textwidth}
\resizebox{\textwidth}{!}{
\begin{tabular}{lc c  c c  c c  c c }
\toprule
\multirow{2}{*}{Methods} & \multicolumn{2}{c}{Java to C\#} & \multicolumn{2}{c}{C\# to Java}  & \multicolumn{2}{c}{Refine Small}  & \multicolumn{2}{c}{Refine Medium}\\ 
\cmidrule(lr){2-3}\cmidrule(lr){4-5}\cmidrule(lr){6-7}\cmidrule(lr){8-9}
& BLEU & EM  & BLEU & EM & BLEU & EM & BLEU & EM  \\ 
\midrule
Naive Copy & 18.54 & 0  & 18.69 & 0 & 78.06 & 0 & 90.91 & 0 \\
RoBERTa (code)  & 77.46 & 56.10  & 71.99 & 57.90 & 77.30 & 15.90 & 90.07 & 4.10 \\
CodeBERT & 79.92 & 59.00  & 72.14 & 58.80 & 77.42 &16.40 &91.07 &5.20 \\ 
GraphCodeBERT & 80.58 & 59.40  & 72.64 & 58.80 & \textbf{80.02} & 17.30 & \textbf{91.31} & 9.10 \\
PLBART  &  83.02 & 64.60  &  78.35 &  65.00 & 77.02 & 19.21 & 88.50& {8.98}\\
\midrule
CodeT5-small & 82.98	& 64.10 &79.10&65.60& 76.23&	19.06 &89.20 &	10.92\\
~~~+dual-gen & 82.24 & 	63.20 & 78.10 &	63.40 & 77.03 &	17.50 & 88.99 &  10.28\\
~~~+multi-task &83.49&64.30&78.56&65.40&77.03&20.94&87.51&11.11\\
\cdashline{1-9}
CodeT5-base &\textbf{84.03}	&\textbf{65.90} &\textbf{79.87}	&\textbf{66.90} &77.43	&21.61 & 87.64	& 13.96 \\
~~~+dual-gen & 81.84 & 62.00 &77.83	& 63.20 & 77.66 &	19.43 & 90.43 & 11.69 \\
~~~+multi-task & 82.31 & 63.40 & 78.01 & 64.00 & 78.06 & \textbf{22.59} & 88.90 & \textbf{14.18} \\
	
\bottomrule
\end{tabular}
}
\captionsetup{type=table}
\vspace{-0.5em}  
\caption{
BLEU-4 scores and exact match (EM) accuracies for code translation (Java to C\# and C\# to Java) and code refinement (small and medium) tasks.
}
\label{table:code_translation}

\end{minipage}

&
\begin{minipage}{0.36\textwidth}
\resizebox{\textwidth}{!}{

\begin{tabular}{lcc}
\toprule
\multirow{2}{*}{Methods} & Defect &  Clone  \\ 
& Accuracy & F1 \\
\midrule
RoBERTa  & 61.05   &94.9 \\
CodeBERT & 62.08  &96.5\\
DOBF & - & 96.5 \\
GraphCodeBERT & -  &97.1 \\ 
PLBART &  63.18 &\textbf{97.2} \\
\midrule
CodeT5-small &63.40&97.1 \\
~~~+dual-gen & 63.47&97.0\\
~~~+multi-task & 63.58 &-\\
\cdashline{1-3}
CodeT5-base &\textbf{65.78}&\textbf{97.2} \\
~~~+dual-gen & 62.88  &97.0\\
~~~+multi-task & 65.02 &-\\

\bottomrule
\end{tabular}
}
\captionsetup{type=table}
\vspace{-0.5em}  
\caption{Results on the code defect detection  and clone detection  tasks.}
\label{table:classification}

\end{minipage}
\end{tabular}
}
\end{center}
\vspace{-2em}
\end{table*}

%% file: sections/exp-res.tex
\section{Results and Analysis}
In this section, we compare CodeT5 with SOTA models on a broad set of CodeXGLUE downstream tasks (\cref{sec:downstream}), and  investigate the effects of our  bimodal dual generation and multi-task learning (\cref{sec:dual_gen_multi_task}), followed by a detailed analysis on the proposed identifier-aware  pre-training (\cref{sec:identifier}).
\subsection{CodeXGLUE Downstream Tasks}\label{sec:downstream}
We evaluate two sizes of our model: CodeT5-small and CodeT5-base that are pre-trained with  identifier-aware denoising. In addition, we consider the model that continues to train with bimodal dual generation (dual-gen) and show the results with   multi-task fine-tuning. The results of all comparison models are obtained from their original papers and also the CodeXGLUE paper~\cite{DBLP:journals/corr/abs-2102-04664}.

\vspace{-0.5em}
\paragraph{Code Summarization.}
We show code summarization results of smoothed BLEU-4 on six PL data in Table~\ref{table:summarize}. We observe all our model variants significantly  outperform prior work with either an encode-only (RoBERTa, CodeBERT, DOBF) or encoder-decoder framework (PLBART). 
Moreover, the salient performance gap between these two groups of models confirms that encode-only frameworks are suboptimal for generation tasks.
Compared to the SOTA encoder-decoder model PLBART, we  find that even our CodeT5-small yields better overall scores (also on Python and Java) given that our model is much smaller (60M vs. 140M) and  PLBART  is pre-trained with  much larger Python and Java data (> 100 times).
We attribute such improvement to our identifier-aware denoising pre-training and better employment of bimodal training data\footnote{Apart from bimodal dual generation, we concatenate NL and PL for training while PLBART deals with them separately.}.
By increasing the model size, our CodeT5-base boosts the overall performance by over 1.2 absolute points over PLBART.

\vspace{-0.5em}
\paragraph{Code Generation.} 
We compare CodeT5 with GPT-style models and PLBART in Table~\ref{table:concode}. 
Our CodeT5-small outperforms all decoder-only models and also the SOTA PLBART, which again confirms the superiority of encoder-decoder models at generating code snippets. Moreover, our CodeT5-base further significantly pushes the SOTA results across three metrics. Particularly, it achieves   around 4.7 points improvement on CodeBLEU over PLBART, indicating  our CodeT5 can better comprehend the code syntax and semantics with the help of identifier-aware pre-training.

\vspace{-0.5em}
\paragraph{Code-to-Code Generation Tasks.} 
We compare two  code-to-code generation tasks: code translation and code refinement in Table~\ref{table:code_translation} and further consider one naive copy baseline by copying the source input as the target prediction.
In the code translation task, our CodeT5-small outperforms most of baselines and obtains comparable results with PLBART, which shows the advantages of encoder-decoder models in the code-to-code generation setting. Our CodeT5-base further achieves consistent  improvements over PLBART across various metrics for translating from Java to C\# and vice versa.

Here we show   one CodeT5's output of translating C\# to Java in Figure~\ref{fig:translate_case}.
In this case, despite the poor BLEU score,  CodeT5 is able to generate a function that reserves the same functionality and even has better readability compared to the ground-truth.
This reveals that  CodeT5 has a good generalization ability instead of memorizing and repeating what it has seen before.
On the other hand, it also  
suggests that BLEU score is not a perfect evaluation metric for code generation tasks, where sometimes  a higher score can instead reflect the problematic copy issues of neural models.

Another code-to-code generation task is code refinement, a challenging task  that requires to detect which parts of code are buggy and fix them via generating a bug-free code sequence.
Due to the large overlap of source and target code, even the naive copy approach yields very high BLEU scores but zero exact matches. Therefore, we focus on the exact match (EM) metric to evaluate on this task. As shown in Table~\ref{table:code_translation}, we observe that EM scores for the small data are consistently higher than the medium one, indicating that it is harder to fix bugs for a longer code snippet. Our CodeT5-base significantly outperforms  all baselines on EM and especially boosts over 4.8 points for the more challenging medium  task (13.96 vs. GraphCodeBERT's 9.10), reflecting its strong code understanding capability.

\input{figures/translate_case}
\paragraph{Understanding Tasks.}
We compare with two understanding tasks of defect detection and clone detection in Table~\ref{table:classification}. 
Specifically, we generate the binary labels as a unigram sequence from the decoder for the defect detection task, while for the clone detection task,  we first obtain the sequence embedding  of each code snippet using the last decoder state following~\citet{DBLP:conf/acl/LewisLGGMLSZ20} and then predict the labels by measuring their similarity.
Both  CodeT5-small and CodeT5-base outperform all baselines on the defect detection task while CodeT5-base yields  2.6 accuracy score improvement than PLBART. 
For the clone detection task, our CodeT5 models achieve comparable results to the SOTA GraphCodeBERT and PLBART models. 
These results demonstrate that with an encode-decoder framework, our CodeT5 can still be adapted well for understanding tasks.

\input{tables/ablation}

\subsection{Effects of Bimodal Dual Generation and Multi-task Learning} \label{sec:dual_gen_multi_task}
We  examine the effects of bimodal dual generation at pre-training and multi-task learning at  fine-tuning. The bimodal pre-training brings consistent improvements for code summarization and generation tasks on both CodeT5-small and CodeT5-base. 
However, this pre-training task does not help and even sometimes slightly hurts the performance for  PL-PL generation  and understanding tasks. We anticipate this is because bimodal dual generation learns a better alignment between PL and NL that naturally benefits the former tasks involving both PL and NL.
As a side effect, this objective could  bias the model towards the PL-NL tasks and affect its performance on PL-PL tasks.

In multi-task learning, it generally improves most of downstream tasks except the code translation and defect detection.
Particularly, it largely boosts the performance on code summarization, which is not surprising as code summarization  takes up the largest portion of sub tasks (six out of thirteen) and thereby benefit the most from the multi-task learning. 
Besides, we observe that multi-task learning consistently improves the performance of code refinement, which might benefit from the joint training of both small and medium refinement data.
Another possible reason is that  multi-task training with defect detection would enable the model to better comprehend the code semantics for bug detection, which is also a necessary intermediate step for  code refinement.

\subsection{Analyzing Identifier-aware  Pre-training}\label{sec:identifier}
We provide an ablation study to examine the contribution of each component in our identifier-aware objective.
Specifically, we compare the performance of our CodeT5-small on four selected tasks  by ablating each of the three objectives: masked span prediction (MSP), identifier tagging (IT), and masked identifier prediction (MIP).
As shown in Table~\ref{table:ablation}, we observe that generally removing one of the objectives would reduce the performance for all tasks, indicating that all objectives contribute to the better code understanding of our CodeT5.
However, the effect of each objective differs across tasks. Specifically,
removing   MSP  would largely reduce the performance of all generation tasks but instead increase the  defect detection performance.
This shows that masked span prediction is more crucial for capturing syntactic information for generation tasks.
On the contrary, removing MIP  would hurt the defect detection task the most, indicating that it might focus more on code semantic understanding.
By combining these objectives, our CodeT5 can better capture both syntactic and semantic information from code.

\input{figures/generation_case}

We further provide outputs from CodeT5 and its variant without MIP and IT on code generation in Figure~\ref{fig:generation_case}.
We observe that  CodeT5 can correctly generate the exact function, while the model without MIP and IT  fails to recover the identifiers of ``s2'' and ``hasField''. This shows our identifier-aware denoising pre-training can better distinguish and leverage the identifier information.

\input{tables/IP_S2S_comp}

We also investigate the identifier tagging performance and find it achieves  over 99\% F1 for all PLs, showing that our CodeT5 can confidently distinguish  identifiers in code. We then check whether MSP and MIP tasks would have conflicts as they employ the same sentinel tokens for masking.
In identifier masking, all occurrences of one unique identifier are replaced with the same sentinel token, resulting in a many-to-one mapping compared to the one-to-one mapping in span prediction.
We compare  models pre-trained with either MSP or MIP, and both on these two tasks in Table~\ref{table:IP_S2S}. 
We report the prediction accuracy and also the ratio of how often they can generate the same number of predictions as the sentinel tokens. We observe that pre-training only with either MIP or MSP would bias the model towards that task, achieving poor accuracy and higher mismatch in number of predictions when applied to the other task.
Interestingly, we find that MIP-only objective can better recover the correct number of predictions in the MSP task than MSP-only does for the MIP task, meaning that it is easier to adapt from   many-to-one mapping  to  one-to-one mapping and difficult for the opposite.
At last, combining them can help our model to make a good trade-off on both tasks.

%% file: figures/translate_case.tex
\begin{figure}[t]
\center
\begin{adjustbox}{width=0.98\linewidth}
\begin{tabular}{lp{8.5cm}}

\toprule
Type & \multicolumn{1}{c}{Code}  \\
\midrule
Target  &
\begin{minipage}{\textwidth}
\begin{minted}[escapeinside=||,fontsize=\small]{java}
public long ramBytesUsed(){
    return BASE_RAM_BYTES_USED+((index!=null)?
           index.ramBytesUsed() : 0);}
  \end{minted}
\end{minipage}\\

\midrule
CodeT5  &
\begin{minipage}{\textwidth}
\begin{minted}[escapeinside=||,fontsize=\small]{java}
public long ramBytesUsed(){
    long sizeInBytes = BASE_RAM_BYTES_USED;
    if (index != null){
        sizeInBytes += index.ramBytesUsed();}
    return sizeInBytes;}
  \end{minted} 
\end{minipage}\\
\bottomrule
\end{tabular}
\end{adjustbox}
\vspace{-0.5em}
\caption{
\small One  translation  (C\# to Java) example  that is semantically correct but with a 50.23\% BLEU-4 score.
}
\label{fig:translate_case}
\vspace{-1.5em}
\end{figure}

%% file: tables/ablation.tex
\begin{table}[t]
\centering
\resizebox{\linewidth}{!}{%
\begin{tabular}{l c c c  c}
\toprule
\multirow{2}{*}{Methods} & Sum-PY & Code-Gen  & Refine Small  & Defect \\
& (BLEU) & (CodeBLEU) & (EM) & (Acc) \\
\midrule
CodeT5 & 20.04 & 41.39 &  19.06 & 63.40 \\
~~~-MSP & 18.93 & 37.44 &15.92& 64.02 \\
~~~-IT & 19.73 &39.21 &18.65 & 63.29 \\
~~~-MIP & 19.81 & 38.25 &18.32& 62.92 \\

\bottomrule
\end{tabular}
}
\vspace{-0.5em}
\caption{Ablation study with CodeT5-small on four selected tasks. ``Sum-PY'' denotes   code summarization  on Python and ``Code-Gen'' denotes  code generation.}

\label{table:ablation}
\vspace{-1em}
\end{table}

%% file: figures/generation_case.tex
\begin{figure}[t]
\center
\begin{adjustbox}{width=0.48\textwidth}
\begin{tabular}{p{2cm}p{8cm}}

\toprule
Type & \multicolumn{1}{c}{Code}  \\
\midrule
Source &
\begin{minipage}{\textwidth}
\begin{minted}[escapeinside=||,fontsize=\small]{text}
|\textbf{Text:}| |returns the string value of the  |
specified field. the value is obtained from  
|whichever scan contains the field|.
|\textbf{Env:}| Scan |\colorbox{pink}{s1}|; Scan |\colorbox{yellow}{s2}|; boolean |\colorbox{pastel_blue}{hasField}|
  \end{minted}
\end{minipage}\\
\midrule
CodeT5  &
\begin{minipage}{\textwidth}
\begin{minted}[escapeinside=||,fontsize=\small]{java}
String function (String arg0){ 
    if (|\colorbox{pink}{s1}|.|\colorbox{pastel_blue}{hasField}|(arg0)) 
        return |\colorbox{pink}{s1}|.|getString|(arg0); 
    else return |\colorbox{yellow}{s2}|.|getString|(arg0);}
  \end{minted} 
\end{minipage}\\
\midrule
W/o MIP+IT &
\begin{minipage}{\textwidth}
\begin{minted}[escapeinside=||,fontsize=\small]{java}
String function (String arg0){ 
    return |\colorbox{pink}{s1}|.|getString|(arg0);}
  \end{minted} 
\end{minipage}\\
\bottomrule
\end{tabular}
\end{adjustbox}
\vspace{-0.5em}
\caption{\label{fig:generation_case}
One code generation example on Concode test set, where our CodeT5 gives a correct prediction. The important identifiers  are highlighted.
}
\vspace{-0.5em}
\end{figure}

%% file: tables/IP_S2S_comp.tex
\begin{table}[t]
\centering
\resizebox{0.97\linewidth}{!}{%

\begin{tabular}{l c c c  c}
\toprule
\multirow{2}{*}{Methods} & \multicolumn{2}{c}{MSP} & \multicolumn{2}{c}{MIP} \\
\cmidrule(lr){2-3} \cmidrule(lr){4-5}
& Acc & \#Pred M   & Acc & \#Pred M  \\
\midrule
MSP-only & 50.13 & 99.80 &2.94& 1.60 \\
MIP-only & 1.68 & 82.40 &  42.75 & 98.80 \\
MIP+MSP & 48.26& 99.60 &42.72 & 98.60 \\

\bottomrule
\end{tabular}
}
\vspace{-1mm}
\caption{Compare MSP and MIP  on a subset of Java in CodeSearchNet.  ``\#Pred M'' denotes the ratio of prediction numbers that matches the sentinel token numbers.}

\label{table:IP_S2S}
\vspace{-1mm}
\end{table}

%% file: sections/conclusion.tex
\section{Conclusion}

We have presented CodeT5, a pre-trained encoder-decoder model that incorporates the token type information from code. We propose a novel identifier-aware pre-training objective to better leverage the identifiers and propose a bimodal dual generation task to learn a better NL-PL alignment using code and its comments. 
Our unified model can support both code understanding and generation tasks and allow for multi-task learning.
Experiments show that CodeT5 significantly outperforms all prior work in most CodeXGLUE tasks. 
Further analysis also reveals its better code comprehension capability across various programming languages.

%% file: sections/impact.tex
\section*{Broader Impact and Ethical Consideration}

Our work generally belongs to NLP applications for software intelligence. With the goal of improving the development productivity of software with machine learning methods, software intelligence research has attracted increasing attention in both academia and industries over the last decade.
Software code intelligence techniques can help developers to reduce tedious repetitive workloads, enhance the programming quality and improve the overall software development productivity. This would considerably decrease their working time and also could potentially reduce the computation and operational cost, as a bug might degrade the system performance or even crash the entire system. 
Our work addresses the fundamental challenge of software code pre-training, our study covers a wide range of code intelligence applications in the software development lifecycle, and the proposed CodeT5 method achieves the state-of-the-art performance on many of the benchmark tasks, showing its great potential benefit towards this goal.

We further discuss the ethical consideration of training CodeT5 and the potential risks when applying it into real-world downstream applications:

\paragraph{Dataset bias.} The training datasets in our study are source code including user-written comments from open source Github repositories and publicly available, which do not tie to any specific application. However, it is possible that these datasets would encode some stereotypes like race and gender from the text comments or even from the source code such as  variables, function and class names. As such,  social biases would be intrinsically embedded into the models trained on them.
As suggested by~\citet{DBLP:journals/corr/abs-2107-03374}, interventions such as filtration or modulation of generated outputs may help to mitigate these biases in code corpus.

\paragraph{Computational cost.} Our model pre-training requires non-trivial computational resources though we have tried our best to carefully design our experiments and improve experiments to save unnecessary computation costs. In fact, compared to the recent large-scale language model Codex~\cite{DBLP:journals/corr/abs-2107-03374}, our CodeT5-base has a much smaller model size of 220M than theirs of 12B ($\sim55\times$).  In addition, we experiment on Google Cloud Platform which purchases carbon credits to reduce its carbon footprint, \eg training CodeT5-base produced around 49.25 kg CO\textsubscript{2} which was totally offset by the provider.
Furthermore, we release our pre-trained models publicly to avoid repeated training for the code intelligence research community.

\paragraph{Automation bias.} As CodeT5 can be deployed to provide coding assistance such as code generation for aiding developers,  automation bias of machine learning systems should be carefully considered, especially for developers who tend to over-rely on the model-generated outputs. Sometimes these systems might produce functions that superficially appear correct but do not actually align with the developer’s intents. If developers unintentionally adopt these incorrect code suggestions, it might cause them much longer time on debugging and even lead to some significant safety issues. We suggest 
practitioners using CodeT5 should always bear in mind that its generation outputs  should be only taken as references which require domain experts for further correctness and security checking.

\paragraph{Security implications.} 
We train CodeT5 on existing  code corpus including CodeSearchNet~\cite{DBLP:journals/corr/abs-1909-09436} and a small fraction of Google BigQuery, both of which are originally collected from public Github repositories.
Pre-trained models might encode some sensitive information (\eg personal addresses or  identification numbers) from the training data. 
Though we have  conducted multi-rounds of data cleaning to mitigate this before training our models, it is still possible that some sensitive information cannot be completely removed. Besides, due to the non-deterministic nature of generation models like CodeT5, it might produce some vulnerable code to harmfully affect the software and even be able to benefit more advanced malware development when deliberately misused.

%% file: sections/ack.tex
\section*{Acknowledgements}
We thank Akhilesh Deepak Gotmare, Amrita Saha, Junnan Li, and  Chen Xing for valuable discussions.
We thank Kathy Baxter for the ethical review.
We also thank our anonymous reviewers for their insightful feedback on our paper.